\documentclass{article} 
\usepackage{iclr2023_conference,times}

\usepackage{amsmath,amsfonts,bm}

\def\eqref#1{equation~\ref{#1}}

\def\1{\bm{1}}

\DeclareMathAlphabet{\mathsfit}{\encodingdefault}{\sfdefault}{m}{sl}
\SetMathAlphabet{\mathsfit}{bold}{\encodingdefault}{\sfdefault}{bx}{n}

\usepackage{hyperref}
\usepackage{graphicx}

\usepackage{booktabs}
\usepackage{url}
\usepackage{xspace}
\usepackage{comment}

\title{AngOFA: Leveraging OFA Embedding \\ Initialization and Synthetic Data for \\Angolan Language Model}

\author{Osvaldo Luamba Quinjica \\
Masakhane NLP \\
\texttt{aosiwaduo@outlook.com} \\
\And
David Ifeoluwa Adelani  \\
Department of Computer Science\\
University College London \\
United Kingdom \\
\texttt{d.adelani@ucl.ac.uk} 
}

\newcommand*{\angbert}{\textsc{AngXLM-R \xspace}}
\newcommand*{\angofa}{\textsc{AngOFA \xspace}}

\iclrfinalcopy 
\begin{document}

\maketitle

\begin{abstract}
In recent years, the development of pre-trained language models (PLMs) has gained momentum, showcasing their capacity to transcend linguistic barriers and facilitate knowledge transfer across diverse languages. However, this progress has predominantly bypassed the inclusion of very-low resource languages, creating a notable void in the multilingual landscape. This paper addresses this gap by introducing four tailored PLMs specifically finetuned for Angolan languages, employing a Multilingual Adaptive Fine-tuning (MAFT) approach. In this paper, we survey the role of informed embedding initialization and synthetic data in enhancing the performance of MAFT models in downstream tasks. We improve baseline over SOTA AfroXLMR-base (developed through MAFT) and OFA (an effective embedding initialization) by 12.3 and 3.8 points respectively.

\end{abstract}



\section{Introduction}



Significant advancements have marked the progress of language models and evaluation datasets across various global languages~\citep{devlin-etal-2019-bert, conneau-etal-2020-unsupervised, workshop2023bloom, xue-etal-2021-mt5}. Nevertheless, this progress has often bypassed numerous African languages, creating a significant gap. Simultaneously, the majority of African-centric language models have overlooked the inclusion of Angolan languages~\citep{dossou-etal-2022-afrolm, alabi-etal-2022-adapting, ogueji-etal-2021-small}. Efforts within the AfricaNLP community have been commendable in broadening downstream evaluation datasets~\citep{adelani-etal-2021-masakhaner, adelani-etal-2022-masakhaner, muhammad-etal-2023-semeval, ma2023taxi1500}. However, despite these initiatives, Angolan languages still lack representation.


 In the pursuit of developing a multilingual pre-trained language model (PLM), there are two primary approaches. The first entails building a model from scratch, training it directly on multiple languages, employing a specific self-supervised learning such as masked language modeling~\citep{devlin-etal-2019-bert}. An alternative approach is multilingual adaptive fine-tuning (MAFT) which involves adapting an existing multilingual pretrained language model to a new set of languages~\citep{alabi-etal-2022-adapting, wang-etal-2022-expanding, imanigooghari-etal-2023-glot500}. MAFT gains favor for its resource efficiency, especially in scenarios where computational budgets pose constraints amid the escalating model sizes~\citep{tay2022scale, gupta2023continual}. The performance of MAFT can be further enhanced by introducing new vocabulary tokens for the additional languages and employing non-Gaussian embedding initialization~\citep{minixhofer-etal-2022-wechsel, dobler-de-melo-2023-focus, liu2023ofa}. 
 
 In this paper, we introduce the first set of multilingual PLMs tailored for five Angolan languages using the MAFT approach. We compare PLMs developed through MAFT with and without informed embedding initialization, denoted as \angofa and \angbert, respectively. Leveraging OFA approach to perform embedding initialization before performing MAFT, our findings reveal that \angofa significantly outperforms \angbert and OFA, underscoring the substantial performance improvements achievable through the incorporation of informed embedding initialization and synthetic data. 

\section{Angolan Languages}


Boasting a rich linguistic landscape comprising more than 40 languages and a population of 32 million people, 
Angolan languages include Portuguese, some Khoisan languages, and mostly Bantu languages from the Niger-Congo family. Despite this linguistic diversity, there is a notable scarcity of literature, radio, or television programming in native Angolan languages. All languages in Angola are written using the Latin script, and many share common digraphs. Due to data scarcity, our focus will primarily revolve around the five most spoken Angolan languages: Umbundu, Kimbundu, Kikongo, Chokwe, and Luba-Kasai. See Table \ref{table-angola-languages} for more details.


\begin{table}[t]
\begin{center}
 \resizebox{\columnwidth}{!}{%
\begin{tabular}{lrr|rrrr}
\toprule
& {\bf Bantu} & {\bf No.} & {\bf NLLB} & {\bf Synthetic} &  {\bf Combined} & {\bf Combined} \\

{\bf Language} & {\bf Zone} & {\bf Speakers} & {\bf  Corpus (MB)} & {\bf Corpus (MB)} & {\bf Corpus (MB)} & {\bf No. Sentences} \\

\midrule
Chokwe (cjk) & Zone K & 0.5M& 11.3 & 108.2 & 119.5 & 878,824  \\
Kimbundu (kmb) & Zone H & 1.7M & 10 & 98.5 & 108.5  & 800,603\\
Kikongo (kon) & Zone H & 2M & 112.1 & 107.9 & 220 & 2,189,413 \\
Luba-Kasai (lua) & Zone L &  0.06M & 133.2 & 113.9 & 247.1 & 2,415,794\\
Umbundu (umb) & Zone R & 6M & 15.1 & 98. 5 & 113.6 & 902,961\\
\midrule
Total & & 10.2M & 281.6 & 527 & 808.6 & 7,187,595\\

\bottomrule
\end{tabular}
}
\vspace{-3mm}
\caption{\textbf{Language Information and Statistics}: Summary of language, language family, number of speakers, number of sentences. All languages belong to the Niger-Congo/Bantu family, we state the Bantu Zones according to \citep{Smith_1949}. Synthetic corpus was generated using NLLB-600M machine translation model}
\label{table-angola-languages}
\end{center}
\end{table}

\section{Approaches to improve MAFT }

\subsection{Vocabulary Expansion}
\label{vocab-expansion}

PLMs are proned to Out-of-Vocabulary (OOV) tokens for languages or scripts uncovered during pre-training. The situation is more pronounced for unseen scripts~\citep{adelani-etal-2021-masakhaner,pfeiffer-etal-2021-unks}, one of the most effective way of dealing with this is to expand the vocabulary of the PLM to cover new tokens~\citep{wang-etal-2019-improving}. Glot-500~\citep{imanigooghari-etal-2023-glot500} was created by first expanding the vocabulary of XLM-R from 250K to 400K before MAFT. However, the new tokens added were randomly initialized.

\subsection{OFA: Embedding Factorization}


OFA addresses two problems of adapting PLMs to new languages (1) the random initialization of embeddings for new subwords fails to exploit the lexical knowledge encoded in the source model (2) the introduction of additional parameters poses potential obstacles to the efficient training of the finetuned model~\citep{liu2023ofa}. OFA solves these problems by leveraging both external multilingual embeddings and embeddings in the source PLM to initialize the embeddings of new subwords.  In its approach, OFA 
factorizes the embeddings matrix of the source PLM into two smaller matrices as replacements. 
Within a lower-dimensional space, the embeddings of non-overlapping new subwords are expressed as combinations of source PLM subword embeddings. These combinations are weighted by similarities derived from well-aligned external multilingual embeddings, i.e., ColexNet+~\citep{liu2023crosslingual}, covering more than one thousand languages. Overlapping subword embeddings are directly copied. This approach ensures that embeddings for subwords shared between the source PLM and the extended vocabulary are integrated, preserving continuity in representation. To complete the process, OFA duplicates all non-embedding parameters from the source PLM model, and the source tokenizer is substituted with the target tokenizer post-vocabulary extension. 



\subsection{Synthetic data for language modeling}
\label{synthetic-data}

For languages lacking sufficient pre-training data, synthetic data can be generated through dictionary augmentation~\citep{reid-etal-2021-afromt} or machine translation (MT) model---an approach very popular in MT research known as back-translation is an effective way to improve MT model for low-resource languages~\citep{sugiyama-yoshinaga-2019-data,xia-etal-2019-generalized}. In this paper, we utilize synthetic data obtained through machine translation as described in ~\citep{adelani2023sib200}. The authors generated machine-translated data for 34 African languages(including Angolan languages) with less than 10MB of data, using the English news commentary dataset ~\citep{kocmi-etal-2022-findings}, which contains over 600K sentences.

\section{Data}

\subsection{Training data}
\label{train_data}

We leveraged the NLLB dataset~\citep{nllb2022}, excluding English translations, and focused solely on Kimbundu, Umbundu, Kikongo, Chokwe, and Luba-Kasai. These languages were concatenated into a single file as our pre-training corpus. 
Additionally, we added synthetic data generated through NLLB. \autoref{table-angola-languages} shows the details of the monolingual data.

\subsection{Evaluation data}
\label{eval_data}
In our work, we evaluated on SIB-200~\citep{adelani2023sib200}, a text classification dataset that provides train/dev/test sets with 7 classes in more than 200 African languages and dialects. The distribution of the classes are:  science/technology (252), travel (198), politics (146), sports (122), health (110), entertainment (93), geography (83).  SIB-200 is the only benchmark dataset that covers Angolan languages.  We evaluated only on the subset of Angolan languages covered in this work.

\section{Experimental Setup}

We utilized the cross-lingual capabilities of XLM-R~\citep{conneau-etal-2020-unsupervised} for training, resulting in the creation of a novel set of PLMs\footnote{Models available at \url{https://github.com/zuela-ai/ANGOFA}}: \angbert and \textsc{AngOFA}. These models, underwent distinct fine-tuning processes. Specifically, \angbert underwent fine-tuning using the MAFT approach outlined in ~\cite{alabi-etal-2022-adapting}, with two variants---one trained solely on monolingual data (281.6 MB), and the other incorporating both monolingual and synthetic data (808.7 MB).  

Similarly, \angofa also underwent two variations of fine-tuning, utilizing the datasets in the same manner as \angbert. However, \angofa followed the configurations outlined for \texttt{ofa-multi-768}, as described in ~\citep{liu2023ofa}.
We opted to maintain 768 as the only latent dimension in our experiments based on insights from~\citep{imanigooghari-etal-2023-glot500, liu2023ofa} and further supported by preliminary results from our own experiments. These findings revealed evidence of information loss in lower dimensions, particularly noticeable in tasks such as text classification. This dataset partitioning approach aimed to investigate the effects of the MAFT and OFA approaches, both with and without synthetic data, on model performance.

We compared our new models to the following baseline models:




\begin{enumerate}
    \item XLM-R~\citep{conneau-etal-2020-unsupervised}: an encoder-only model that underwent pre-training on 100 languages through a masked language model objective. XLM-R does not cover any language evaluated in this work. 
    \item Serengeti~\citep{adebara-etal-2023-serengeti}: trained on 500 African languages, including 10 high-resource ones. It includes Kimbundu, Umbundu, and Chokwe.
    \item Glot-500~\citep{imanigooghari-etal-2023-glot500}:  derived from XLM-R, was extended to cover 500 languages by expanding its vocabulary from 250K to 400K, thus accommodating new tokens representing 400 languages previously absent in XLM-R. Glot-500 covers all Angolan languages used in our evaluation. 
    \item AfroXLMR-base~\citep{alabi-etal-2022-adapting}: developed using the MAFT approach, it covers 20 languages with a monolingual corpus of at least 50MB. Angolan languages are not included.
    \item AfroXLMR-base-76L~\citep{adelani2023sib200}: developed using the MAFT approach, it covers languages with at least 10MB of data on the web. It expands coverage to include more languages, notably those listed in the NLLB-200 MT model. Synthetic data was also generated for approximately 30 languages with limited data, including all five Angolan languages. In total, it covers 76 languages. 
    \item OFA~\citep{liu2023ofa}: integrates OFA embedding initialization alongside MAFT using Glot500-c~\citep{imanigooghari-etal-2023-glot500}, thus including all languages addressed in this work. 
\end{enumerate}


\section{Results and Discussion}

\begin{table}[t]
\begin{center}
 \resizebox{\columnwidth}{!}{%
\begin{tabular}{lrr|rrr|rr|rr|r}
\toprule
 & \multicolumn{2}{c}{\textit{Pre-trained (scratch)}}  & \multicolumn{3}{c}{\textit{MAFT}}  & \multicolumn{2}{c}{\textit{MAFT} + syn. data}  & \multicolumn{2}{c}{\textit{OFA}} &  \textit{OFA + syn} \\

 &  & & {\bf Glot}  & {\bf Afro} &  {\bf \textsc{Ang}} & {\bf Afro} & {\bf \textsc{Ang}} & {\bf \textsc{Ang}} & {\bf OFA}\\
 
{\bf Lang. } & {\bf XLM-R} & {\bf Serengeti} & {\bf 500}  & {\bf XLMR}  & {\bf \textsc{XLM-R}} & {\bf XLMR76} & {\bf \textsc{XLM-R}} & {\bf \textsc{OFA}} & {\bf 500} & {\bf \textsc{AngOFA}}\\
\midrule
cjk & 41.3 & 43.2 & 42.9 & 51.3  & 43.6 & 55.6 & 51.7  & 46.3 & 52.8 & \bf 58.4 \\
kmb & 44.8 & 46.9 & 43.5 & 50.6 & 50.2 & 58.5 & 56.6  &  58.5 & 63.2 & \bf 64.7 \\
kon & 67.8 & 69.1 & 72.6 & 65.7 & 72.5 & 77.2 & 76.1  & 78.8 & 76.9 & \bf 82.4    \\
lua & 54.5 & 57.9 & 54.7 & 62.5 & 65.4 & 64.4 & 73.2 & 69.1 & 68.6  & \bf 73.5\\
umb & 50.4 & 51.7  & 40.3 & 50.5 & 54.9 & 61.0 & 56.8 & 54.3 & 61.8 & \bf 63.3  \\
\hline
\textbf{Ave.} & 51.8 &53.7 & 50.7 & 56.1 & 57.3 & 63.3 & 62.8 & 61.4 & 64.6 & \bf 68.4 \\
\hline
\end{tabular}
}
\vspace{-3mm}
\caption{\textbf{Benchmark results}: comparing the effectiveness of OFA to random initialization before multilingual adaptive fine-tuning (MAFT)}
\label{table-1}
\end{center}
\end{table}


Table\ref{table-1} shows the performance of our baseline models using the \textbf{weighted F1 metric}. We discuss our key findings below: 

\paragraph{Region-specific PLMs are better than those pre-trained from scratch with many languages} Our results shows that \angbert created with MAFT performed better than XLM-R, AfroXLMR, Serengeti and Glot-500 with $+5.5$, $+1.2$, $+3.6$, $+6.6$ points respectively. The last two PLMs have been pre-trained on 500+ languages with few Angolan languages but performed worse than AfroXLMR (adapted through MAFT to 20 languages), and \angbert (adapted to five Angolan languages). This shows that region-specific PLMs covering related languages within the same language family can be more effective. 

\paragraph{MAFT results can be boosted by leveraging synthetic monolingual data} By incorporating additional synthetic data, \angbert (+SYN data) performance improved by $+5.5$ over the \angbert without synthetic data. However, it failed to beat the performance of AfroXLMR-base-76L that has been trained on 76 African languages including all Angolan languages except for Luba-Kasai with the largest data. Our experiment showed that the adapted PLM to 76 languages performed better than Serengeti pre-trained on 500 languages,  which further shows that we can create better PLMs to cover more languages through adaptation without the expensive process of pre-training from scratch.

\paragraph{OFA embedding initialization with larger data is more effective} 
Models initialized with OFA demonstrated a consistent improvement compared with other baselines. This indicates that OFA, which explicitly leverages information encoded in the embeddings of the source model and external multilingual embeddings, is superior to random initialization. Notably, \textsc{AngOFA}'s advantage over OFA is accentuated by its access to a significantly larger corpus of data for the respective languages through the use of synthetic data. Without the additional synthetic data \angofa performed worse than \textsc{OFA} pre-trained on 500 languages with a drop of $-3.2$. However, when we trained on the synthetic data, \angofa achieved the best overall performance with $+16.6$ over XLM-R, $+12.3$ over AfroXLMR, and $+5.6$ over \angbert (with synthetic data).

\section{Conclusion and Future work}
This paper introduces four multilingual PLMs models tailored for Angolan languages. Our experimental findings illustrate that employing informed embedding initialization significantly enhances the performance of a MAFT model in downstream tasks. While models initialized with OFA exhibit superior results compared to their counterparts, even in the case where \angbert finetuned on a larger corpus of data for the respective languages performs poorly as compared to OFA finetuned on a smaller corpus. Nevertheless, the specific factors contributing to \angbert's superiority over OFA, especially in the context of Luba-Kassai, raise intriguing questions about the primary determinants influencing the performance of models in downstream tasks, including considerations like dataset size versus informed embedding initialization. These questions are left for future investigation. Furthermore, we aim to expand the application of OFA to more African languages for further exploration.


\subsubsection*{Acknowledgments}
This work was supported in part by Oracle Cloud credits and related resources provided by Oracle. David Adelani acknowledges the support of DeepMind Academic Fellowship programme.

\bibliography{iclr2023_conference}

@inproceedings{ogueji-etal-2021-small,
    title = "Small Data? No Problem! Exploring the Viability of Pretrained Multilingual Language Models for Low-resourced Languages",
    author = "Ogueji, Kelechi  and
      Zhu, Yuxin  and
      Lin, Jimmy",
    editor = "Ataman, Duygu  and
      Birch, Alexandra  and
      Conneau, Alexis  and
      Firat, Orhan  and
      Ruder, Sebastian  and
      Sahin, Gozde Gul",
    booktitle = "Proceedings of the 1st Workshop on Multilingual Representation Learning",
    month = nov,
    year = "2021",
    address = "Punta Cana, Dominican Republic",
    publisher = "Association for Computational Linguistics",
    url = "https://aclanthology.org/2021.mrl-1.11",
    doi = "10.18653/v1/2021.mrl-1.11",
    pages = "116--126",
    
}

@inproceedings{dossou-etal-2022-afrolm,
    title = "{A}fro{LM}: A Self-Active Learning-based Multilingual Pretrained Language Model for 23 {A}frican Languages",
    author = "Dossou, Bonaventure F. P.  and
      Tonja, Atnafu Lambebo  and
      Yousuf, Oreen  and
      Osei, Salomey  and
      Oppong, Abigail  and
      Shode, Iyanuoluwa  and
      Awoyomi, Oluwabusayo Olufunke  and
      Emezue, Chris",
    editor = {Fan, Angela  and
      Gurevych, Iryna  and
      Hou, Yufang  and
      Kozareva, Zornitsa  and
      Luccioni, Sasha  and
      Sadat Moosavi, Nafise  and
      Ravi, Sujith  and
      Kim, Gyuwan  and
      Schwartz, Roy  and
      R{\"u}ckl{\'e}, Andreas},
    booktitle = "Proceedings of The Third Workshop on Simple and Efficient Natural Language Processing (SustaiNLP)",
    month = dec,
    year = "2022",
    address = "Abu Dhabi, United Arab Emirates (Hybrid)",
    publisher = "Association for Computational Linguistics",
    url = "https://aclanthology.org/2022.sustainlp-1.11",
    doi = "10.18653/v1/2022.sustainlp-1.11",
    pages = "52--64",
    
}

@inproceedings{alabi-etal-2022-adapting,
    title = "Adapting Pre-trained Language Models to {A}frican Languages via Multilingual Adaptive Fine-Tuning",
    author = "Alabi, Jesujoba O.  and
      Adelani, David Ifeoluwa  and
      Mosbach, Marius  and
      Klakow, Dietrich",
    editor = "Calzolari, Nicoletta  and
      Huang, Chu-Ren  and
      Kim, Hansaem  and
      Pustejovsky, James  and
      Wanner, Leo  and
      Choi, Key-Sun  and
      Ryu, Pum-Mo  and
      Chen, Hsin-Hsi  and
      Donatelli, Lucia  and
      Ji, Heng  and
      Kurohashi, Sadao  and
      Paggio, Patrizia  and
      Xue, Nianwen  and
      Kim, Seokhwan  and
      Hahm, Younggyun  and
      He, Zhong  and
      Lee, Tony Kyungil  and
      Santus, Enrico  and
      Bond, Francis  and
      Na, Seung-Hoon",
    booktitle = "Proceedings of the 29th International Conference on Computational Linguistics",
    month = oct,
    year = "2022",
    address = "Gyeongju, Republic of Korea",
    publisher = "International Committee on Computational Linguistics",
    url = "https://aclanthology.org/2022.coling-1.382",
    pages = "4336--4349",
    
}

@misc{adelani2023sib200,
      title={SIB-200: A Simple, Inclusive, and Big Evaluation Dataset for Topic Classification in 200+ Languages and Dialects}, 
      author={David Ifeoluwa Adelani and Hannah Liu and Xiaoyu Shen and Nikita Vassilyev and Jesujoba O. Alabi and Yanke Mao and Haonan Gao and Annie En-Shiun Lee},
      year={2023},
      eprint={2309.07445},
      archivePrefix={arXiv},
      primaryClass={cs.CL},
      url = "https://arxiv.org/abs/2309.07445"
}

@inproceedings{devlin-etal-2019-bert,
    title = "{BERT}: Pre-training of Deep Bidirectional Transformers for Language Understanding",
    author = "Devlin, Jacob  and
      Chang, Ming-Wei  and
      Lee, Kenton  and
      Toutanova, Kristina",
    editor = "Burstein, Jill  and
      Doran, Christy  and
      Solorio, Thamar",
    booktitle = "Proceedings of the 2019 Conference of the North {A}merican Chapter of the Association for Computational Linguistics: Human Language Technologies, Volume 1 (Long and Short Papers)",
    month = jun,
    year = "2019",
    address = "Minneapolis, Minnesota",
    publisher = "Association for Computational Linguistics",
    url = "https://aclanthology.org/N19-1423",
    doi = "10.18653/v1/N19-1423",
    pages = "4171--4186",
   
}

@inproceedings{conneau-etal-2020-unsupervised,
    title = "Unsupervised Cross-lingual Representation Learning at Scale",
    author = "Conneau, Alexis  and
      Khandelwal, Kartikay  and
      Goyal, Naman  and
      Chaudhary, Vishrav  and
      Wenzek, Guillaume  and
      Guzm{\'a}n, Francisco  and
      Grave, Edouard  and
      Ott, Myle  and
      Zettlemoyer, Luke  and
      Stoyanov, Veselin",
    editor = "Jurafsky, Dan  and
      Chai, Joyce  and
      Schluter, Natalie  and
      Tetreault, Joel",
    booktitle = "Proceedings of the 58th Annual Meeting of the Association for Computational Linguistics",
    month = jul,
    year = "2020",
    address = "Online",
    publisher = "Association for Computational Linguistics",
    url = "https://aclanthology.org/2020.acl-main.747",
    doi = "10.18653/v1/2020.acl-main.747",
    pages = "8440--8451",
    
}

@inproceedings{wang-etal-2022-expanding,
    title = "Expanding Pretrained Models to Thousands More Languages via Lexicon-based Adaptation",
    author = "Wang, Xinyi  and
      Ruder, Sebastian  and
      Neubig, Graham",
    editor = "Muresan, Smaranda  and
      Nakov, Preslav  and
      Villavicencio, Aline",
    booktitle = "Proceedings of the 60th Annual Meeting of the Association for Computational Linguistics (Volume 1: Long Papers)",
    month = may,
    year = "2022",
    address = "Dublin, Ireland",
    publisher = "Association for Computational Linguistics",
    url = "https://aclanthology.org/2022.acl-long.61",
    doi = "10.18653/v1/2022.acl-long.61",
    pages = "863--877",
}

@inproceedings{imanigooghari-etal-2023-glot500,
    title = "Glot500: Scaling Multilingual Corpora and Language Models to 500 Languages",
    author = {ImaniGooghari, Ayyoob  and
      Lin, Peiqin  and
      Kargaran, Amir Hossein  and
      Severini, Silvia  and
      Jalili Sabet, Masoud  and
      Kassner, Nora  and
      Ma, Chunlan  and
      Schmid, Helmut  and
      Martins, Andr{\'e}  and
      Yvon, Fran{\c{c}}ois  and
      Sch{\"u}tze, Hinrich},
    editor = "Rogers, Anna  and
      Boyd-Graber, Jordan  and
      Okazaki, Naoaki",
    booktitle = "Proceedings of the 61st Annual Meeting of the Association for Computational Linguistics (Volume 1: Long Papers)",
    month = jul,
    year = "2023",
    address = "Toronto, Canada",
    publisher = "Association for Computational Linguistics",
    url = "https://aclanthology.org/2023.acl-long.61",
    doi = "10.18653/v1/2023.acl-long.61",
    pages = "1082--1117",
}

@inproceedings{
tay2022scale,
title={Scale Efficiently: Insights from Pretraining and Finetuning Transformers},
author={Yi Tay and Mostafa Dehghani and Jinfeng Rao and William Fedus and Samira Abnar and Hyung Won Chung and Sharan Narang and Dani Yogatama and Ashish Vaswani and Donald Metzler},
booktitle={International Conference on Learning Representations},
year={2022},
url={https://openreview.net/forum?id=f2OYVDyfIB}
}

@inproceedings{
gupta2023continual,
title={Continual Pre-Training of Large Language Models: How to re-warm your model?},
author={Kshitij Gupta and Benjamin Th{\'e}rien and Adam Ibrahim and Mats Leon Richter and Quentin Gregory Anthony and Eugene Belilovsky and Irina Rish and Timoth{\'e}e Lesort},
booktitle={Workshop on Efficient Systems for Foundation Models @ ICML2023},
year={2023},
url={https://openreview.net/forum?id=pg7PUJe0Tl}
}

@misc{liu2023ofa,
      title={OFA: A Framework of Initializing Unseen Subword Embeddings for Efficient Large-scale Multilingual Continued Pretraining}, 
      author={Yihong Liu and Peiqin Lin and Mingyang Wang and Hinrich Schütze},
      year={2023},
      eprint={2311.08849},
      archivePrefix={arXiv},
      primaryClass={cs.CL},
      url="https://arxiv.org/abs/2311.08849"
}

@inproceedings{
liu2023crosslingual,
title={Crosslingual Transfer Learning for Low-Resource Languages Based on Multilingual Colexification Graphs},
author={Yihong Liu and Haotian Ye and Leonie Weissweiler and Renhao Pei and Hinrich Schuetze},
booktitle={The 2023 Conference on Empirical Methods in Natural Language Processing},
year={2023},
url={https://openreview.net/forum?id=Tn5hALAaA4}
}

@article{nllb2022,
  title={No Language Left Behind: Scaling Human-Centered Machine Translation},
  author={NLLB-Team and Marta Ruiz Costa-juss{\`a} and James Cross and Onur cCelebi and Maha Elbayad and Kenneth Heafield and Kevin Heffernan and Elahe Kalbassi and Janice Lam and Daniel Licht and Jean Maillard and Anna Sun and Skyler Wang and Guillaume Wenzek and Alison Youngblood and Bapi Akula and Lo{\"i}c Barrault and Gabriel Mejia Gonzalez and Prangthip Hansanti and John Hoffman and Semarley Jarrett and Kaushik Ram Sadagopan and Dirk Rowe and Shannon L. Spruit and C. Tran and Pierre Yves Andrews and Necip Fazil Ayan and Shruti Bhosale and Sergey Edunov and Angela Fan and Cynthia Gao and Vedanuj Goswami and Francisco Guzm'an and Philipp Koehn and Alexandre Mourachko and Christophe Ropers and Safiyyah Saleem and Holger Schwenk and Jeff Wang},
  journal={ArXiv},
  year={2022},
  volume={abs/2207.04672},
  url={https://api.semanticscholar.org/CorpusID:250425961}
}

@inproceedings{adebara-etal-2023-serengeti,
    title = "{SERENGETI}: Massively Multilingual Language Models for {A}frica",
    author = "Adebara, Ife  and
      Elmadany, AbdelRahim  and
      Abdul-Mageed, Muhammad  and
      Alcoba Inciarte, Alcides",
    editor = "Rogers, Anna  and
      Boyd-Graber, Jordan  and
      Okazaki, Naoaki",
    booktitle = "Findings of the Association for Computational Linguistics: ACL 2023",
    month = jul,
    year = "2023",
    address = "Toronto, Canada",
    publisher = "Association for Computational Linguistics",
    url = "https://aclanthology.org/2023.findings-acl.97",
    doi = "10.18653/v1/2023.findings-acl.97",
    pages = "1498--1537",
}

@article{adelani-etal-2021-masakhaner,
    title = "{M}asakha{NER}: Named Entity Recognition for {A}frican Languages",
    author = "Adelani, David Ifeoluwa  and
      Abbott, Jade  and
      Neubig, Graham  and
      D{'}souza, Daniel  and
      Kreutzer, Julia  and
      Lignos, Constantine  and
      Palen-Michel, Chester  and
      Buzaaba, Happy  and
      Rijhwani, Shruti  and
      Ruder, Sebastian  and
      Mayhew, Stephen  and
      Azime, Israel Abebe  and
      Muhammad, Shamsuddeen H.  and
      Emezue, Chris Chinenye  and
      Nakatumba-Nabende, Joyce  and
      Ogayo, Perez  and
      Anuoluwapo, Aremu  and
      Gitau, Catherine  and
      Mbaye, Derguene  and
      Alabi, Jesujoba  and
      Yimam, Seid Muhie  and
      Gwadabe, Tajuddeen Rabiu  and
      Ezeani, Ignatius  and
      Niyongabo, Rubungo Andre  and
      Mukiibi, Jonathan  and
      Otiende, Verrah  and
      Orife, Iroro  and
      David, Davis  and
      Ngom, Samba  and
      Adewumi, Tosin  and
      Rayson, Paul  and
      Adeyemi, Mofetoluwa  and
      Muriuki, Gerald  and
      Anebi, Emmanuel  and
      Chukwuneke, Chiamaka  and
      Odu, Nkiruka  and
      Wairagala, Eric Peter  and
      Oyerinde, Samuel  and
      Siro, Clemencia  and
      Bateesa, Tobius Saul  and
      Oloyede, Temilola  and
      Wambui, Yvonne  and
      Akinode, Victor  and
      Nabagereka, Deborah  and
      Katusiime, Maurice  and
      Awokoya, Ayodele  and
      MBOUP, Mouhamadane  and
      Gebreyohannes, Dibora  and
      Tilaye, Henok  and
      Nwaike, Kelechi  and
      Wolde, Degaga  and
      Faye, Abdoulaye  and
      Sibanda, Blessing  and
      Ahia, Orevaoghene  and
      Dossou, Bonaventure F. P.  and
      Ogueji, Kelechi  and
      DIOP, Thierno Ibrahima  and
      Diallo, Abdoulaye  and
      Akinfaderin, Adewale  and
      Marengereke, Tendai  and
      Osei, Salomey",
    editor = "Roark, Brian  and
      Nenkova, Ani",
    journal = "Transactions of the Association for Computational Linguistics",
    volume = "9",
    year = "2021",
    address = "Cambridge, MA",
    publisher = "MIT Press",
    url = "https://aclanthology.org/2021.tacl-1.66",
    doi = "10.1162/tacl_a_00416",
    pages = "1116--1131"
}

@inproceedings{adelani-etal-2022-masakhaner,
    title = "{M}asakha{NER} 2.0: {A}frica-centric Transfer Learning for Named Entity Recognition",
    author = "Adelani, David  and
      Neubig, Graham  and
      Ruder, Sebastian  and
      Rijhwani, Shruti  and
      Beukman, Michael  and
      Palen-Michel, Chester  and
      Lignos, Constantine  and
      Alabi, Jesujoba  and
      Muhammad, Shamsuddeen  and
      Nabende, Peter  and
      Dione, Cheikh M. Bamba  and
      Bukula, Andiswa  and
      Mabuya, Rooweither  and
      Dossou, Bonaventure F. P.  and
      Sibanda, Blessing  and
      Buzaaba, Happy  and
      Mukiibi, Jonathan  and
      Kalipe, Godson  and
      Mbaye, Derguene  and
      Taylor, Amelia  and
      Kabore, Fatoumata  and
      Emezue, Chris Chinenye  and
      Aremu, Anuoluwapo  and
      Ogayo, Perez  and
      Gitau, Catherine  and
      Munkoh-Buabeng, Edwin  and
      Memdjokam Koagne, Victoire  and
      Tapo, Allahsera Auguste  and
      Macucwa, Tebogo  and
      Marivate, Vukosi  and
      Elvis, Mboning Tchiaze  and
      Gwadabe, Tajuddeen  and
      Adewumi, Tosin  and
      Ahia, Orevaoghene  and
      Nakatumba-Nabende, Joyce  and
      Mokono, Neo Lerato  and
      Ezeani, Ignatius  and
      Chukwuneke, Chiamaka  and
      Oluwaseun Adeyemi, Mofetoluwa  and
      Hacheme, Gilles Quentin  and
      Abdulmumin, Idris  and
      Ogundepo, Odunayo  and
      Yousuf, Oreen  and
      Moteu, Tatiana  and
      Klakow, Dietrich",
    editor = "Goldberg, Yoav  and
      Kozareva, Zornitsa  and
      Zhang, Yue",
    booktitle = "Proceedings of the 2022 Conference on Empirical Methods in Natural Language Processing",
    month = dec,
    year = "2022",
    address = "Abu Dhabi, United Arab Emirates",
    publisher = "Association for Computational Linguistics",
    url = "https://aclanthology.org/2022.emnlp-main.298",
    doi = "10.18653/v1/2022.emnlp-main.298",
    pages = "4488--4508"
    
}

@inproceedings{muhammad-etal-2023-semeval,
    title = "{S}em{E}val-2023 Task 12: Sentiment Analysis for {A}frican Languages ({A}fri{S}enti-{S}em{E}val)",
    author = "Muhammad, Shamsuddeen Hassan  and
      Abdulmumin, Idris  and
      Yimam, Seid Muhie  and
      Adelani, David Ifeoluwa  and
      Ahmad, Ibrahim Said  and
      Ousidhoum, Nedjma  and
      Ayele, Abinew Ali  and
      Mohammad, Saif  and
      Beloucif, Meriem  and
      Ruder, Sebastian",
    editor = {Ojha, Atul Kr.  and
      Do{\u{g}}ru{\"o}z, A. Seza  and
      Da San Martino, Giovanni  and
      Tayyar Madabushi, Harish  and
      Kumar, Ritesh  and
      Sartori, Elisa},
    booktitle = "Proceedings of the 17th International Workshop on Semantic Evaluation (SemEval-2023)",
    month = jul,
    year = "2023",
    address = "Toronto, Canada",
    publisher = "Association for Computational Linguistics",
    url = "https://aclanthology.org/2023.semeval-1.315",
    doi = "10.18653/v1/2023.semeval-1.315",
    pages = "2319--2337"
}

@misc{ma2023taxi1500,
      title={Taxi1500: A Multilingual Dataset for Text Classification in 1500 Languages}, 
      author={Chunlan Ma and Ayyoob ImaniGooghari and Haotian Ye and Ehsaneddin Asgari and Hinrich Schütze},
      year={2023},
      eprint={2305.08487},
      archivePrefix={arXiv},
      primaryClass={cs.CL}, 
      url = "https://arxiv.org/abs/2305.08487"
}

@inproceedings{xue-etal-2021-mt5,
    title = "m{T}5: A Massively Multilingual Pre-trained Text-to-Text Transformer",
    author = "Xue, Linting  and
      Constant, Noah  and
      Roberts, Adam  and
      Kale, Mihir  and
      Al-Rfou, Rami  and
      Siddhant, Aditya  and
      Barua, Aditya  and
      Raffel, Colin",
    editor = "Toutanova, Kristina  and
      Rumshisky, Anna  and
      Zettlemoyer, Luke  and
      Hakkani-Tur, Dilek  and
      Beltagy, Iz  and
      Bethard, Steven  and
      Cotterell, Ryan  and
      Chakraborty, Tanmoy  and
      Zhou, Yichao",
    booktitle = "Proceedings of the 2021 Conference of the North American Chapter of the Association for Computational Linguistics: Human Language Technologies",
    month = jun,
    year = "2021",
    address = "Online",
    publisher = "Association for Computational Linguistics",
    url = "https://aclanthology.org/2021.naacl-main.41",
    doi = "10.18653/v1/2021.naacl-main.41",
    pages = "483--498"
}

@inproceedings{minixhofer-etal-2022-wechsel,
    title = "{WECHSEL}: Effective initialization of subword embeddings for cross-lingual transfer of monolingual language models",
    author = "Minixhofer, Benjamin  and
      Paischer, Fabian  and
      Rekabsaz, Navid",
    editor = "Carpuat, Marine  and
      de Marneffe, Marie-Catherine  and
      Meza Ruiz, Ivan Vladimir",
    booktitle = "Proceedings of the 2022 Conference of the North American Chapter of the Association for Computational Linguistics: Human Language Technologies",
    month = jul,
    year = "2022",
    address = "Seattle, United States",
    publisher = "Association for Computational Linguistics",
    url = "https://aclanthology.org/2022.naacl-main.293",
    doi = "10.18653/v1/2022.naacl-main.293",
    pages = "3992--4006"
}

@article{Smith_1949,
title={The Classification of the Bantu Languages. By Malcolm Guthrie, Ph.D. Published for the International African Institute by the Oxford University Press, 1948. Pp. 91. Map. 8s. 6d. net.},
volume={19}, 
DOI={10.2307/1156267}, 
number={1},
journal={Africa},
author={Smith, Edwin W.}, 
year={1949}, 
pages={73–74}
}
\bibliographystyle{iclr2023_conference}


\end{document}